\documentclass[runningheads,a4paper]{llncs}

\usepackage{amssymb}
\usepackage{graphicx}

\usepackage{url}

\urldef{\mailsa}\path|ar.ahmadi62@gmail.com, tani1216jp@gmail.com|
\newcommand{\keywords}[1]{\par\addvspace\baselineskip
\noindent\keywordname\enspace\ignorespaces#1}

\usepackage{multirow}
\usepackage{adjustbox}
\usepackage{booktabs}

\begin{document}

\mainmatter  

\title{Bridging the Gap between Probabilistic and Deterministic Models: A Simulation Study on a Variational Bayes Predictive Coding Recurrent Neural Network Model}

\titlerunning{Bridging the Gap between Probabilistic and Deterministic Models}

%
%
\author{Ahmadreza Ahmadi\inst{1,2}%
\and Jun Tani\inst{1,2}
\thanks{Corresponding author.}%
}
\authorrunning{A. Ahmadi and J. Tani}

\institute{Dept. of Electrical Engineering, KAIST, Daejeon, 305-701, Korea\\
\and Okinawa Institute of Science and Technology, Okinawa, Japan 904-0495\\
\mailsa\\
}

%
%

\toctitle{Lecture Notes in Computer Science}
\tocauthor{Authors' Instructions}
\maketitle

\begin{abstract}
The current paper proposes a novel variational Bayes predictive coding RNN model, which can learn to generate fluctuated temporal patterns from exemplars. The model learns to maximize the lower bound of the weighted sum of the regularization and reconstruction error terms. We examined how this weighting can affect development of different types of information processing while learning fluctuated temporal patterns. Simulation results show that strong weighting of the reconstruction term causes the development of deterministic chaos for imitating the randomness observed in target sequences, while strong weighting of the regularization term causes the development of stochastic dynamics imitating probabilistic processes observed in targets. Moreover, results indicate that the most generalized learning emerges between these two extremes. The paper concludes with implications in terms of the underlying neuronal mechanisms for autism spectrum disorder and for free action.
\end{abstract}

\section{Introduction}

Cognitive agents dealing with a changing environment need to develop internal models accounting for such fluctuations by extracting underlying structures through learning. Recently, many schemes have been proposed for learning fluctuated temporal patterns by extracting latent probabilistic structures. Those schemes include conventional dynamic Bayesian networks such as HMM and Kalman filters, and the recently developed variational Bayes recurrent neural network (RNN) models~\cite{bayer2014learning,chung2015recurrent,gregor2015draw}. At the same time, there have been alternative trials with a deterministic dynamics approach in which predictive models for probabilistically generated target sequences are learned by embedding extracted probabilistic structures into deterministic chaos self-organized in RNN models~\cite{namikawa2011neurodynamic,nishimoto2004learning,tani1995embedding}. Because these two approaches - one probabilistic and the other deterministic - have been developed relatively independently, research bridging the gap between them seems worthwhile.

In this direction, Murata et al.~\cite{murata2017learning} developed a predictive coding-type stochastic RNN model inspired by the free energy minimization principle~\cite{friston2005theory}. This model learns to predict the mean and variance of sensory input for each next time step of multiple perceptual sequences, mapping from its current latent state. The learning optimizes not only connectivity weights but also the latent state at the initial step for each training sequence by way of back-propagation through time~\cite{Rumelhart:1986:LIR:104279.104293}. Murata and colleagues experimented with the degree of the initial state dependency in learning to imitate probabilistic sequences, and examined how the internal dynamic structure develops differently in different cases. It turned out that initial state dependency arbitrated the development between deterministic and probabilistic dynamic structure. In the case of strong initial state dependency, deterministic chaos is dominantly developed, while stochastic dynamics develops by estimating larger variance in the case of weak initial state dependency.

However, the model in~\cite{murata2017learning} suffers a considerable drawback. It cannot estimate variance for latent variables (the context units). This significantly constrains the capability of the model in learning latent probabilistic structures in target patterns. The current paper proposes a version of a variational Bayes RNN (VBRNN) model, which can estimate variance for each context unit at each time step (time varying variance). The proposed model is simpler than other VBRNN models~\cite{bayer2014learning,chung2015recurrent,gregor2015draw} because it employs a predictive coding scheme rather than an autoencoder. Thus, the model is referred to as the variational Bayes predictive coding RNN (VBP-RNN). It assumes that learning aims to maximize the lower bound \textit{L}~\cite{Bishop:2006:PRM:1162264}, which is represented by the weighted sum of a negative regularization term for the posterior distribution of the latent variable and the likelihood term in the output generation. 

The current study investigates how differently weighting these two terms in the summation during learning influences the development of different types of information processing in the model. To this end, we conducted a simulation experiment in which the VBP-RNN learned to predict/generate fluctuated temporal patterns containing probabilistic transitions between prototypical patterns. Consistent with Murata et al.~\cite{murata2017learning}, results showed that the different weighting arbitrates between two extremes at which the model develops either a deterministic dynamic structure or a probabilistic one. Analysis on simulation results clarifies how the degree of generalization in learning as well as the strength of the top-down intentionality in generating patterns changes from one extreme to another.

\section{Model}

This section introduces the lower bound equation. Then, the variational Bayes predictive coding RNN (VBP-RNN) is described.

\subsection{Generative Model and the Lower Bound}

A generative model can provide probabilistic prediction about fluctuating sensation. The joint probability of sensation \textit{x} and latent variable \textit{z} in a generative model can be written as a product of likelihood and a prior as:
\begin{equation} \label{eq:1}
    P_\theta(x,z) = P_\theta(x|z)P(z)
\end{equation}
where the likelihood~$P_\theta(x|z)$ is parameterized by learning parameter~$\theta$. On the other hand, perception of \textit{x} can be considered as a process of inferring posterior \textit{z} as~$P(z|x)$ which, however, becomes intractable when the likelihood is a nonlinear function of~$\theta$. Then, the problem is to maximize the joint probability~$P_\theta(x,z)$ for a given sensory dataset~$X = \left\{x_t\right\}_{t=1}^N$ by inferring both~~$\theta$ and the true posterior which is approximated using a recognition model~$Q_\theta(z|x)$. 
In variational Bayes, it has been well known that this approximation by means of minimization of the KL-divergence between the model approximation~\textit{Q} and the true posterior~\textit{P} is equivalent to maximizing a value referred to as the lower bound~\cite{Bishop:2006:PRM:1162264}. The lower bound \textit{L} to be maximized can be written as: 
\begin{equation} \label{eq:2}
    L = -KL(Q_\theta(Z|X)||P(Z)) + E_{Q_{\theta}(Z|X)}[log(P_\theta(X|Z)]  
\end{equation}
where the first term on the right hand side is the regularization term by which the posterior distribution of the latent variable is constrained to be similar to its prior, usually taken as a unit Gaussian distribution. The second term minimizes the reconstruction error. This formula for maximizing the lower bound is equivalent to the principle of free energy minimization provided by Friston~\cite{friston2005theory}.

\subsection{Variational Bayes Predictive Coding RNN Models}
Here, we describe the implementation of the aforementioned formulation in a continuous-time RNN (CTRNN) model as well as in a multiple timescale RNN (MTRNN) model~\cite{Ahmadi20173,yamashita2008emergence}. If we take~~$X^l = \left\{x_t^l\right\}_{t=1}^N$ to be the \textit{l}th teaching sensory sequence pattern used for training the VBP-RNN model, the regularization term in the lower bound for the all teaching sequences~$L_z$ can be written as:
\begin{equation} \label{eq:3}
    L_z = \sum^L_{l=1} -KL(Q_\theta(z^l_{1:T}|x^l_{1:T}) || P(z^l_{1:T})) 
\end{equation}
where the posterior~$z_t^l$ is approximated by the recognition model~$Q_\theta$ as a conditional probability with a given sensory sequence pattern. The prior~$P(z_t^l)$ can be given, for example as a normal distribution. Next, the reconstruction error term~$L_x$ can be described as:
\begin{equation} \label{eq:4}
    L_x = \sum^L_{l=1} [E_{Q_{\theta}(z^l_{1:T}|x^l_{1:T})}\sum^T_{t=1}[log(P_\theta(x^l_{t}|z^l_{t})]]
\end{equation}
The total lower bound is obtained as a weighted sum of the regularization term and the reconstruction error term, with a weighting parameter \textit{W} as shown in Equation~\ref{eq:5}.
\begin{equation} \label{eq:5}
    L = W.L_z + (1-W)L_x
\end{equation}
Finally, the objective of learning is to maximize the total lower bound by optimizing both learning parameter~$\theta$ and~$z_1^l$ as the latent state at the initial step for each latent state sequence, given a specific value for \textit{W}.
Following the Kingma and Welling's reparameterization trick~\cite{kingma2013auto}, a random value is sampled from a standard normal distribution at each time step, i.e.~$\varepsilon^l_t\sim N(0,1)$ which is used to sample~$z^l_{1:T}$ from~\textit{Q}. In the current RNN implementation, the latent state is represented as the ensemble of the internal state values of all context units at each step, as~$z_{t,1:C}^l$. Then, the internal state of the \textit{i}th context unit at time step \textit{t} in the \textit{l}th sequence can be computed with the estimation of time varying mean~$\mu_{t,i}^l$ and variance~$\sigma_{t,i}^l$ in the following way:
\begin{equation} \label{eq:6}
    z^l_{t,i} = \mu^l_{t,i} + \sigma^l_{t,i}\varepsilon^l_{t,i}
\end{equation}
\begin{equation} \label{eq:7}
    \mu^l_{t,i} = (1 - \frac{1}{\tau_i})z^l_{t-1,i} + \frac{1}{\tau_i}(\sum_j w^{\mu c}_{ij}c^l_{t-1,j} + \sum_k w^{\mu x}_{ik}x^l_{t-1,k} + b^\mu_i)     
\end{equation}
\begin{equation} \label{eq:8}   
    \sigma^l_{t,i} = \exp (0.5\times \sum_j (w^{\sigma c}_{ij}c^l_{t-1,j}+b^\sigma_i))
\end{equation}
where~$\tau_i$ is time constant of the \textit{i}th context unit. Although the VBP-CTRNN uses the same time constant value for all context units, the VBP-MTRNN uses different time constant values for different units. Learning parameters~$w_{ij}^{\mu|\sigma}$ and~$b_i^{\mu|\sigma}$ are connectivity weights and biases, respectively.~$c_{t,i}^l$ is the activation of the \textit{i}th context unit the value of which is computed as~$c_{t,i}^l=tanh(z^l_{t,i})$. Because both~$P(z_{1:T}^t)$ (the prior) and~$Q_\theta(z^l_{1:T}|x^l_{1:T})$ are Gaussian, the regularization term in the lower bound for \textit{C} numbers of context units can be rewritten as:
\begin{equation} \label{eq:10}
    L_z = \frac{1}{2C}\sum^L_{l=1}\sum^T_{t=1} \sum^C_{i=1}(1+log((\sigma^l_{t,i})^2)-(\mu^l_{t,i})^2-(\sigma^l_{t,i})^2)
\end{equation}
The \textit{i}th dimension of the prediction output is computed in the form of probabilistic distribution by using the SoftMax function with \textit{M} elements. Eventually, the distribution is computed by mapping from the current context unit activation patterns at each time step. The probability of the \textit{i}th dimension of the prediction output is computed with~$w_{ij}^{xc}$ learnable connectivity weights and~$b_i^x$ biases as:
\begin{equation} \label{eq:9}   
    x^l_{t,i} = \frac{exp(\sum_j (w^{xc}_{ij}c^l_{t,j}+b_i^x))}{\sum_{i=1}^M exp(\sum_j (w^{xc}_{ij}c^l_{t,j}+b_i^x))}    
\end{equation}

\begin{figure}[t!]
\centering
\includegraphics[height=7cm]{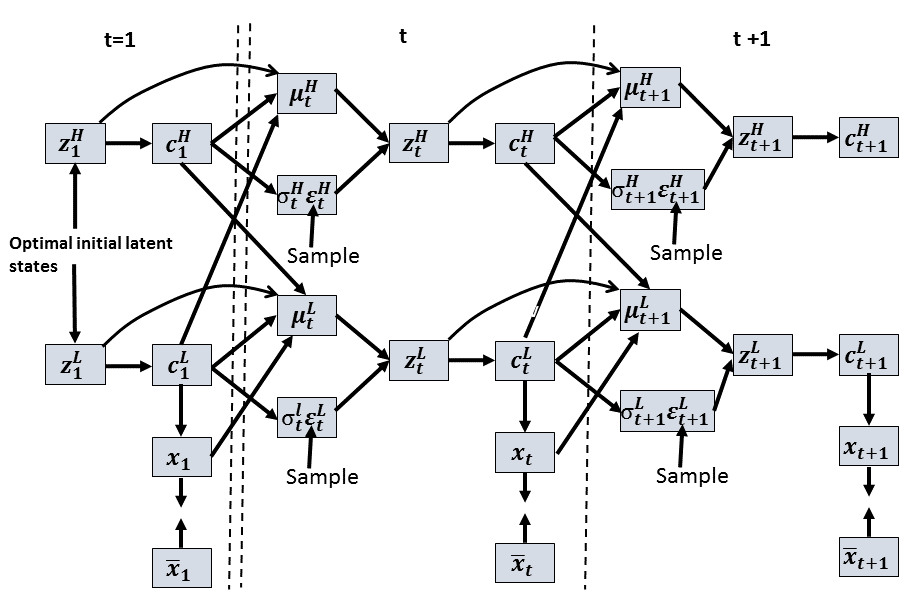}
\caption{The scheme of variational Bayes predictive MTRNN (VBP-MTRNN). \textit{H} and \textit{L} are abbreviation for Higher and Lower layers, respectively. The initial time step at which optimal initial latent states are given to the network is shown by t=1. The word ``Sample" shows that for each context unit, a random value is sampled from a standard normal distribution at each time step ($\varepsilon^{H|L}_t\sim N(0,1)$).}
\label{fig1}
\end{figure}

Figure~\ref{fig1} outlines the information flow in the VBP-MTRNN. The learning process starts with random initialization for all learning parameters and the initial latent state for each latent sequence. The lower bound \textit{L} with a given \textit{W} can be obtained for each epoch by computing the latent state sequences as well as the output sequences using equations 6 to 10. \textit{L} is maximized by optimizing the learning parameters and the initial latent state for each latent sequence by using the back-propagation through time (BPTT) algorithm~\cite{Rumelhart:1986:LIR:104279.104293}. With an optimal \textit{W} at the convergence of training, we expected the model's posterior sequence to approximate the true one . 
This model is simple compared to other variational Bayes RNN models~\cite{bayer2014learning,chung2015recurrent,gregor2015draw}. Those models are built from separate functions, of the decoder RNN and the encoder RNN. By optimizing the connectivity weights and the initial latent states, in the current model the same RNN computes both the prediction output sequences by means of its forward dynamics and the posterior of the latent state sequences by means of BPTT.

\section{Simulation Experiments}

\begin{figure}[t!]
\centering
\includegraphics[height=5.6cm]{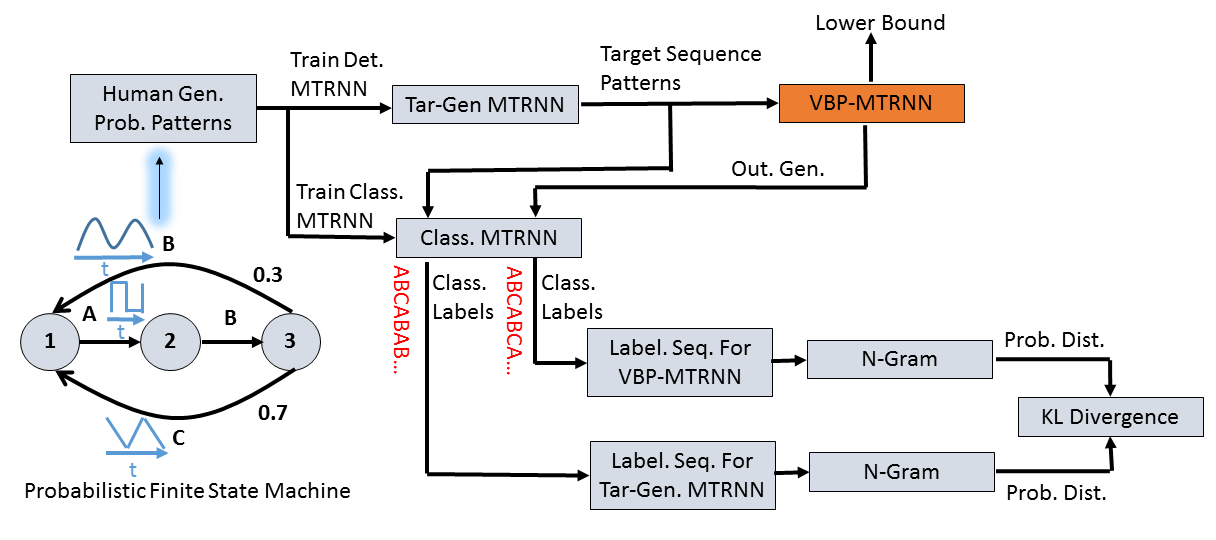}
\caption{Illustration of the simulation and analysis procedure. Human Gen. Prob. Patterns, Tar-Gen MTRNN, Class. MTRNN, Class. Labels, Label. Seq., and Prob. Dist. are abbreviations for Human-Generated Probabilistic Patterns, Target Generator MTRNN, Classifier MTRNN, Classified Labels, Label Sequences, and Probabilistic Distribution, respectively.}
\label{fig2}
\end{figure}

We conducted simulation experiments to determine how learning in the proposed model depends on meta-prior \textit{W} as well as on the number of training patterns. Figure~\ref{fig2} illustrates simulation and analysis performed in this study. First, a human generated patterns like ABB ABC ABC ABB ABC in which after AB, B is 30$\%$ probable C is 70$\%$. These probability transitions are the same as those of a probabilistic finite state machine (pFSM) as shown in the bottom left of Figure~\ref{fig2}. Then, we trained two types of MTRNNs, a target generator MTRNN (Tar-Gen-MTRNN) and an output classifier MTRNN (Class-MTRNN) with those prototypical patterns. The Tar-Gen-MTRNN was prepared for autonomous generation of target temporal patterns (consisting of 100000 steps) which were then used for training the VBP-MTRNN. The Class-MTRNN was prepared for autonomous segmentation of temporal patterns into sequences of labels assigned to different prototypical patterns, which were used for the N-GRAM analysis. The patterns generated by the Tar-Gen-MTRNN were used as the target teaching patterns for the main experiment, training the VBP-MTRNN under different conditions. After training, the characteristics of output patterns generated by the VBP-MTRNN were quantitatively compared with those of the Tar-Gen-MTRNN. Using the trained Class-MTRNN, we computed the probabilistic distribution of \textit{N} consecutive labels corresponding to different prototypical patterns classified from output patterns generated for 100000 steps by both the Tar-Gen-MTRNN and the VBP-MTRNN. Finally, the N-GRAM KL-divergence between these two probability distributions was computed in order to obtain a measure of similarity in the output generation between the Tar-Gen-MTRNN and the VBP-MTRNN trained in different conditions.

All network models consisted of 7 context layers with (from lowest to highest layer) 121, 60, 30, 15, 10, 10, 10 context units, and all were trained for 100,000 epochs. The time constants of the 7 context layers were set to 2, 4, 8, 16, 32, 64, and 128 from lowest to highest layer. There were 121 input and output units in each network. In order to map the target patterns from 2-D to 121-D, the 2-D SoftMax transform was used (11$\times$11). All VBP-MTRNN models used a mini-batch size equal to 8, and the ADAM optimizer~\cite{kingma2014adam} was used to maximize the weighted lower bound.

\subsection{Target Generator MTRNN and Classifier MTRNN}
The Tar-Gen-MTRNN was prepared for autonomous generation of fluctuating target patterns of 100000 steps, which were used for the main experiment training the VBP-MTRNN. The 2-D temporal patterns for training the Tar-Gen-MTRNN were provided by a human generate pattern compositions from a set of different prototypical patterns using a tablet input device. The target sequence pattern was generated by concatenating 30 prototypical patterns. Each prototypical pattern is a different periodic pattern with 3 cycles, fluctuating in amplitude and periodicity at each appearance. The prototypical pattern ``A" for example is generated 10 times in training patterns by a human. Because it is generated by a human on a tablet, each prototypical pattern is expressed with different amplitudes and periodicity in each trial. After training with these human expressed patterns, the Tar-Gen-MTRNN generated an output sequence pattern for 100,000 steps. This output sequence pattern was generated closed-loop (feeding next step inputs with current step prediction outputs) while adding Gaussian noise with zero mean and with constant~$\sigma$ of 0.1 into the internal state of each context unit at each time step. This was done in order to make the network output patterns stochastic while maintaining a certain probabilistic structure in transitions between the prototypical patterns. We sampled the sequence pattern generated from the 50,000th step to the 100,000th step. Then, two groups of target patterns were sampled, one consisting of 16 sequence patterns each with 400 step length and the other of 128 sequence patterns with the same step length. These target sequence patterns were used in the main experiment, training the VBP-MTRNN.

In order to prepare the Class-MTRNN, an MTRNN model was trained to classify 3 different prototypical patterns (A, B, and C). The same human-generated sequence pattern consisting of 30 consecutive prototypical patterns was used as the teaching input pattern. The corresponding label sequence pattern was used as the target of SoftMax output with 3 elements for A, B, and C labels.

\subsection{Simulation Experiment of VBP-MTRNN and Analysis}\label{VBP-MTRNN}

\begin{figure}[t!]
\centering
\includegraphics[height=9cm]{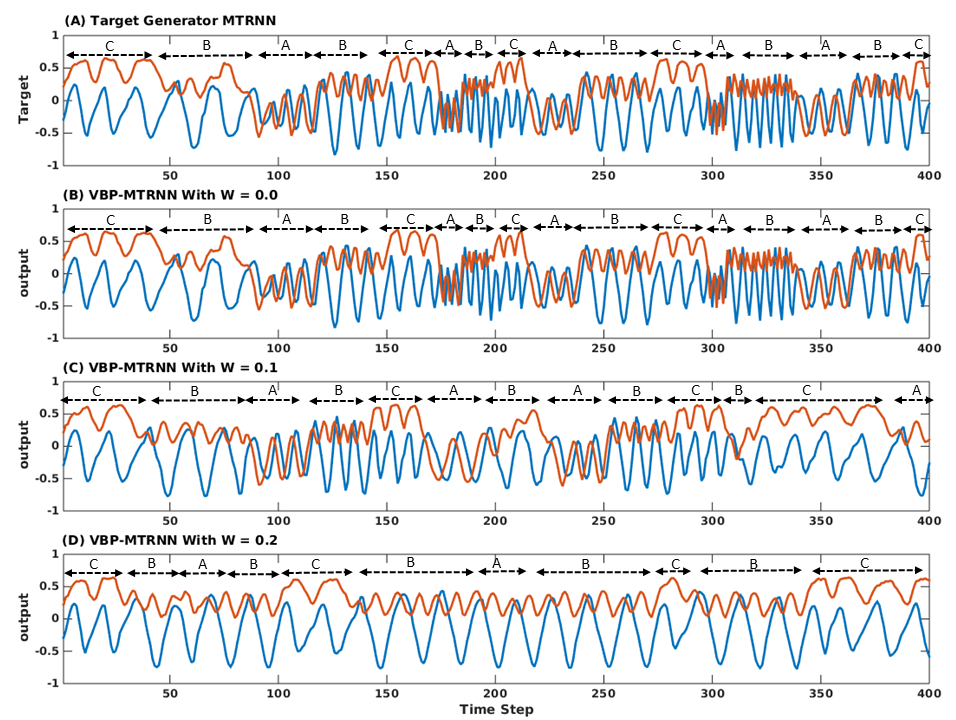}
\caption{A typical comparison between a teaching target sequence pattern and the closed-loop output generated by the VBP-MTRNN model trained with 16 target sequences and set with different values for \textit{W}. The capital letters segmenting the sequence patterns indicate label sequences as classified by the Class-MTRNN.}
\label{fig3}
\end{figure}

The VBP-MTRNN trained with meta-prior \textit{W} set to 0.0, 0.01, 0.1, and 0.2 and with both 16 and 128 teaching target sequences that had been generated by the Tar-Gen-MTRNN. After training for 100,000 epochs under each condition, closed-loop output patterns were generated starting from all different initial latent states obtained for all target sequences after learning. Figure~\ref{fig3} compares one target sequence pattern and its corresponding closed-loop regeneration by the VBP-MTRNN trained with 16 target sequences. The first row shows the target pattern and the second, the third and the fourth rows show regenerated patterns with \textit{W} set to 0.0, 0.1, and 0.2, respectively. Each pattern is associated with a sequence of labels classified by the Class-MTRNN. When \textit{W} = 0.0, the target sequence pattern was completely regenerated for all steps. When \textit{W} = 0.1, target and the regenerated patterns begin to significantly diverge at around 170 steps. Local deviation from each prototypical pattern arose soon after onset. When \textit{W}=0.2, the divergence starts earlier. 

These observations suggest that the VBP-MTRNN trained with \textit{W}=0.0 develops deterministic dynamics. Additional analysis on the output sequence generated for 100,000 steps revealed that there was no periodicity, suggesting that deterministic chaos or transient chaos developed in this learning condition. On the other hand, when \textit{W} was increased to larger values during training, the model developed probabilistic processing in which more randomness was generated internally.

\begin{figure}[t!]
\centering
\includegraphics[height=9cm]{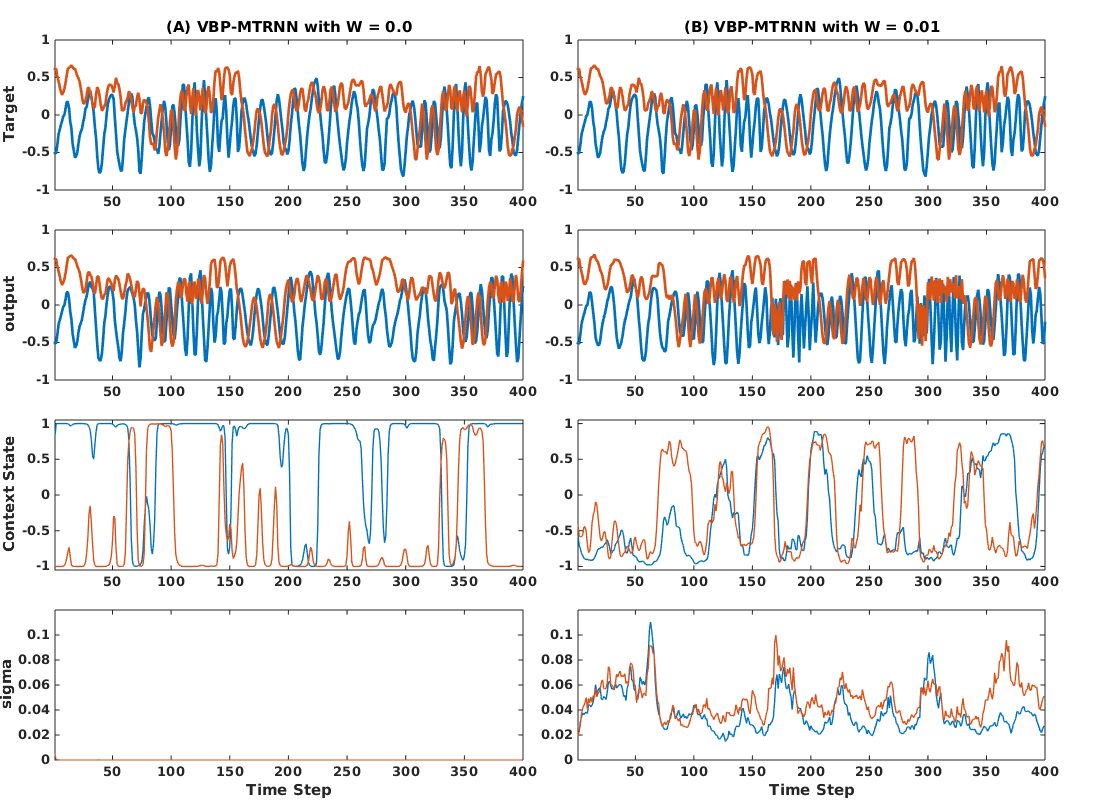}
\caption{The target, generated output, context state, and sigma values of two VBP-MTRNNs trained with 128 target sequences and meta-prior \textit{W} set to (A) 0.0 and (B) 0.01.}
\label{fig4}
\end{figure}

The sigma and context state values of two selected neural units inside the second context layer for VBP-MTRNNs trained with 128 target sequences and with \textit{W} equal to 0.0 and 0.01 are given in Figure~\ref{fig4}. From first to fourth rows are one of the target sequence patterns, its corresponding closed-loop regeneration, the context states, and the sigma values. With \textit{W} set to 0.0, the sigma values become close to zero accounting for the development of deterministic dynamics. With \textit{W} set to 0.01, the sigma values fluctuate from 0.0 and 0.1 accounting for the development of stochastic dynamics. Summarily, changing the meta-prior affects sigma values as the VBP-MTRNN alters between deterministic and stochastic dynamics.

\begin{table}[]
    \caption{Average divergence step (ADS) and Tri-gram KL-divergence of Tar-Gen-MTRNN and VBP-MTRNN trained under various conditions.}    
    \centering    
    \begin{adjustbox}{max width=\textwidth}
    \begin{tabular}{c| c| c c c c}    
    \cline{1-3} 
    \toprule
           \multicolumn{2}{c|}{} & \multicolumn{4}{c}{Weighting Parameters} \\

           \multicolumn{2}{r|}{No. Seq.}    & W=0.0   & W=0.01  & W=0.1 & W=0.2\\             
    \hline
            \multirow{2}{*}{ADS} & 16 Seq.  & 370  & 182 & 77 & 50\\

            & 128 Seq. &207 &169 &75 &41\\ 
    \hline        
            \multirow{2}{*}{Tri-gram KL-Div.} & 16 Seq.  & 0.0699  & 0.01 & 0.016 & 0.0817\\
  
            & 128 Seq. &0.015 &0.0151 &0.0033 &0.0575\\ 
        
    \bottomrule
    \end{tabular}
    \end{adjustbox}
    \label{tab1}
\end{table} 
In order to quantify differences in network characteristics when trained under different meta-prior \textit{W} settings and different numbers of training sequence patterns, the average divergence step (ADS) and the N-GRAM KL-divergence between the Tar-Gen-MTRNN and the VBP-MTRNN were computed for each condition. The ADS was computed for all target sequences by taking the average step at which the target sequence pattern and the regenerated one diverged. Starting from the initial step for both 16 and 128 target sequences, divergence was detected when the mean square error between the target and the generated pattern exceeded a threshold (0.025 in the current experiment). N-GRAM KL-divergence can be obtained as described previously. Setting N=3, the Tri-gram was computed for the Tar-Gen-MTRNN. For this purpose, a sample label sequence was generated by feeding the Class-MTRNN with a sequence pattern generated for 100,000 steps by the Tar-Gen-MTRNN. In the same way, a Tri-gram for the VBP-MTRNN which had been trained with each different condition was computed by generating a sample sequence pattern with the same step length. Table~\ref{tab1} shows the results of this analysis. As expected, for both 16 and 128 target sequences, ADS decreases as \textit{W} increases.

The ADS obtained for the 128 target case decreases as the \textit{W} value increases, but not as much overall as does the 16 target case. This suggests that the top-down intention for regenerating a learned sequence pattern decays as \textit{W} and number of training sequences increases. The Tri-gram KL-divergence is minimized in between the two extreme \textit{W} settings, 0.0 and 0.2. This value is smaller for 128 targets than for 16 for each \textit{W} value (except for \textit{W}=0.01, which is very close). The training condition with 128 training target sequences and \textit{W} set to 0.1 turned out to generate the minimum KL-divergence. Interestingly, we found that the probability distribution of the Tri-grams generated by the Tar-Gen-MTRNN and the VBP-MTRNN become quite similar in this condition. Good generalization in learning can be achieved in such a condition by extracting precise probabilistic structures.

It should be taken into account that the average divergence step (ADS) was computed with a time step length of 400 for each target teaching sequence pattern, and that generated patterns exhibited the same time step length, so a higher ADS value does not signify better generalization capability. It is a proper measure only of exact similarity between the target teaching patterns and the generated output patterns. However, each Tri-gram KL-divergence value was computed by comparing long sequence patterns (100,000 steps) test-generated by the Tar-Gen-MTRNN and by the trained VBP-MTRNN. This value represents the capability for generalization in learning, as it shows how much each VBP-MTRNN model was able to extract of the probabilistic structure latent in the teaching target patterns.

\section{Discussion and Summary}

The current paper proposed a novel variational Bayes predictive coding RNN model, which can learn to predict/generate fluctuated temporal patterns from exemplars in order to shed new light on the gap between deterministic and probabilistic modeling. The model network is characterized by a meta-prior \textit{W} which balances two cost functions, the regularization term and reconstruction error term. We investigated how this meta-prior parameter along with the number of taught sequence patterns affects model learning performance through simulation experiments which involved learning sequence patterns which exhibit probabilistic transitions between prototypical patterns. Results are summarized as follows. 

When the meta-prior \textit{W} was set to 0.0, the model learned to imitate the probabilistic transitions observed in taught sequences through the development of deterministic chaos. It was able to repeat taught sequences exactly for long periods. However, with strong top-down intentionality developing in initial latent states, generalization in learning turned out to be poor. When \textit{W} was set with larger values, the model exhibited stochastic dynamics as it adapted time varying sigma to noise sampling. With \textit{W} set too high, reconstruction error increased with additional randomness in generated patterns as taught sequences could be repeated only for shorter periods. For every value of \textit{W}, when the number of taught patterns increased, generalization in learning improved. Moreover, the highest degree of generalization in learning is exhibited in the middle between extreme \textit{W} values. Related to this, one may ask if there are any means to optimize \textit{W} values automatically. It is speculated that \textit{W} values at each layer could be optimized independently by feeding back the system performance in terms of generalization such as by using reinforcement learning. This challenge is left for future study. The current study did not involve with simulation experiments on active inference~\cite{friston2005theory} of the latent state. The active inference can be done by back-propagating the error signal to the context units in the past window~\cite{Ahmadi20173,murata2017learning} for real-time update of their activation values in the direction of minimizing the error. It is expected that smaller value being set for \textit{W}, more effective the active inference becomes because of stronger causality between the latent state and the perceived sensation.  Otherwise, the active inference becomes less effective because of less causality. This expectation is analogous to~\cite{murata2017learning}. 

The current results bear on the task of learning to extract latent structures in observed fluctuating temporal patterns, and as such may inform inquiry into the mechanism underlying autism spectrum disorders (ASD). Van de Cruys et al.~\cite{van2014precise} have suggested that ASD might be caused by overly strong top-down prior potentiation to minimize prediction error, which can enhance capacities for rote learning while losing the capacity to generalize what is learned, a pathology typical of ASD. With the meta-prior \textit{W} set below a threshold value, the proposed model naturally reflects such pathology. This unbalance setting of \textit{W} might be caused by Dysfuction in GABA signaling as shown in mice experiments conducted by Chao and colleagues~\cite{chao2010dysfunction}. Future study is expected to validate this hypothesis. Another implication of the current study involves mechanisms underlying spontaneous or free generation of action~\cite{haggard2008human,libet1985unconscious,tani2016exploring}. Sequences of "seemingly" freely selecting each action can be accounted for by either deterministic chaos with unconscious intentionality, or by stochastic dynamics driven by sampled noise without strong intentionality, or it could be by an intermediate between these two extremes as shown in the current paper. 

Future studies should concern scaling of the proposed model in various real world applications including robot learning, which will inevitably involve dealing with fluctuating temporal patterns. This should include an investigation for the optimization scheme for the meta-prior \textit{W} at each level by feeding back the system performance, as mentioned previously. At the same time, studies should explore the organizing principles of cognitive brains both in normal and abnormal conditions by selectively extending the model, and comparing model with empirical data. Finally, study should aim at understanding how the qualitative meaning of free action differs when generated under differing conditions, important when considering attribution of responsibility for human action for example.
\bibliography{mybib}

\end{document}